\documentclass[12pt]{article}

\usepackage[a4paper,left=2.5cm,right=2cm,top=3cm,bottom=3cm]{geometry}
\usepackage{amssymb}
\usepackage{amsthm}

\usepackage{lineno,hyperref}

\usepackage{subfig}
\usepackage{varwidth} 

\usepackage{graphicx}
\usepackage{mathtools}
\usepackage{xcolor}
\usepackage[normalem]{ulem}

\title{Backward Gradient Normalization in Deep Neural Networks}

\begin{document}
\maketitle

\begin{center}
\author{Alejandro Cabana and Luis F. Lago-Fernández}\\[0.5cm]
\author{Escuela Politécnica Superior, 
  Universidad Autónoma de Madrid,\\
  28049 Madrid, Spain}\\[0.5cm]
\author{alejandro.cabana@estudiante.uam.es, luis.lago@uam.es}
\end{center}

\begin{abstract}
We introduce a new technique for gradient normalization during neural network training. The gradients are rescaled during the backward pass using normalization layers introduced at certain points within the network architecture. These normalization nodes do not affect forward activity propagation, but modify backpropagation equations to permit a well-scaled gradient flow that reaches the deepest network layers without experimenting vanishing or explosion. Results on tests with very deep neural networks show that the new technique can do an effective control of the gradient norm, allowing the update of weights in the deepest layers and improving network accuracy on several experimental conditions.
\end{abstract}



\section{Introduction}
\label{sec:introduction}

Deep Neural Networks (DNNs) are machine learning algorithms with many processing layers between the input and the output \cite{DBLP:journals/nature/LeCunBH15,DBLP:journals/nn/Schmidhuber15}. They are hierarchically organized to learn features of increasing complexity using gradient backpropagation. In the last decade, DNNs have dominated the field with many successful stories in computer vision, natural language processing and many other domains, with models of continuously increasing depth. 
However, proper training of very deep neural networks can be difficult in practice due to the {\em vanishing} and {\em exploding} gradients problems \cite{DBLP:journals/tnn/BengioSF94,DBLP:journals/ijufks/Hochreiter98,DBLP:conf/icml/PascanuMB13}. The vanishing gradients problem refers to a situation where the gradient information can not reach the deepest layers of a multilayer network. In this situation the weights in the initial layers can not be updated, and the network displays slow learning rates and premature convergence. The exploding gradients problem refers to the converse situation, where the gradients increase without bound as they flow backward through the network, leading to unstable learning. 

Several solutions to deal with the vanishing and the exploding gradients problems have been explored in the literature. A simple fix to alleviate the effects of exploding gradients is to clip the gradient to a constant norm \cite{DBLP:conf/icml/PascanuMB13}. For the vanishing gradients problem, solutions include weight initialization techniques, specially designed network architectures, unsupervised layer-wise pre-training, better activation functions and different forms of normalization. A good weight initialization can accelerate convergence and improve learning in DNNs. Several tricks have been proposed that scale the initial weights according to the number of inputs (fan-in) and outputs (fan-out) of the neurons \cite{DBLP:series/lncs/Bengio12}. Weight scaling ensures that the input to the activation function lies initially in the region where the derivative is maximum, hence facilitating the gradient flow. The same effect can be obtained with batch normalization \cite{DBLP:conf/icml/IoffeS15}, where the neuron's preactivation is normalized, using the actual batch data, before application of the non-linearity. Other forms of normalization aimed at accelerating learning in deep networks include layer normalization \cite{DBLP:journals/corr/BaKH16} and weight normalization \cite{DBLP:conf/nips/SalimansK16}. 

The use of network architectures specially designed to cope with the vanishing gradients problem is also a common approach. Long short-term memory (LSTM) networks \cite{DBLP:journals/neco/HochreiterS97} are perhaps the most paradigmatic example. Unlike regular RNNs, where the gradients flow through a succession of matrix products and non-linearities, in an LSTM unit the gradients follow a path where they remain almost unchanged and can continuously feed the errors back. A similar idea is used in Residual Neural Networks \cite{DBLP:conf/cvpr/HeZRS16}, very deep convolutional networks which include skip connections that jump over some layers. These shortcuts allow the gradients to reach the deepest network layers following a more direct path.
Units with rectifier activations such as the ReLU \cite{DBLP:conf/icml/NairH10} can also improve learning in deep neural networks. Due to a constant gradient in their active region, they suffer less from the vanishing gradients problem, and usually outperform saturating units such as the logistic sigmoid or the hyperbolic tangent activation functions \cite{DBLP:journals/jmlr/GlorotBB11}. The ReLU has become the default activation in most deep learning applications. 

In this article we introduce a new approach to the problem that fits within normalization techniques. The method, which we call Backward Gradient Normalization (BGN), rescales the gradients at certain points during backpropagation, guaranteeing that the error signal flows through the network with an almost constant norm and reaches even the deepest layers without much alteration. Our approach is similar to the layer-wise gradient normalization method in \cite{DBLP:journals/corr/YuLSC17}, and inspired by Normalized Gradient Descent (NGD) methods \cite{DBLP:conf/nips/HazanLS15,DBLP:journals/corr/Levy16a}. The  NGD algorithm uses only the gradient direction, but not its magnitude, to update the network weights during training. This can provide numerical stability and improve convergence. Intuitively, by constraining the gradient norm, gradient vanishing and explosion are more easily avoided. The idea behind the block-normalization of Yu et al. \cite{DBLP:journals/corr/YuLSC17} is similar, but the gradient is normalized separately for each layer or block. The main problem with their approach is that the complete gradient must be computed using the backpropagation algorithm before it is normalized layer-wise. Hence, if the network is suffering from the vanishing gradients problem, the gradients computation may run into numerical issues that can not be corrected with normalization. 

Our proposal is to normalize the gradients during, not after, backpropagation. We include several normalization nodes within the network that rescale the gradient during the backward pass. It is this normalized gradient which is sent back to previous layers, guaranteeing that the whole network receives a well-scaled gradient signal. Our weight update equations are formally equivalent to the layer-wise gradient normalization in \cite{DBLP:journals/corr/YuLSC17}, but with the additional advantage that we avoid computing the whole gradient before normalization.

We run several experiments with networks of increasing depth using the permutation invariant MNIST problem. Our results show that: (i) the backward normalized gradients are proportional to the original gradients, with a constant that can be different for each layer; (ii) the normalized gradients are able to reach the deepest layers without degradation; (iii) when BGN is introduced there is more adaptation of the weights in the initial layers; and (iv) inclusion of gradient normalization always improves the results regardless of the base method. Additionally, BGN is less time consuming than other popular normalization techniques such as batch normalization. 

The rest of the article is organized as follows. In section \ref{sec:gradnorm} we introduce the BGN technique and analyze how it affects the gradients backpropagation. In section \ref{sec:gradient_visualization} we present some experimental results that illustrate how the gradients flow through the network under several conditions. Section \ref{sec:experiments} describes the main experimental settings, and section \ref{sec:results} presents the results of our experiments, comparing the accuracy of networks of different depth trained with and without batch normalization and BGN.
Finally, section \ref{sec:conclusions} contains the conclusions and discussion.

\section{Backward Gradient Normalization}
\label{sec:gradnorm}

In the context of training a neural network with a method based on gradients, each layer has two different behaviors. On the forward pass, they receive an input, compute the weighted sum and return the result of the activation function. The concatenation of these operations from the first to the last layer yields the network's output for a given input. On the backward pass, each layer receives a gradient with respect to its output and returns the gradient with respect to the previous layer's output, which enables the use of backpropagation as a method to calculate the gradient of the loss function with respect to the weights and biases of every layer.

Let us take as an example a standard dense layer with the identity function as its activation (or effectively, no activation at all). In the forward pass, it computes the output $\bf z$ as the product of its weights $\bf W$ and its input $\bf x$, plus its bias $\bf b$, as $\bf z = \bf W \bf x + \bf b$. In the backward pass, however, the layer receives the gradient of the loss function $L$ with respect to the layer's output $\nabla_{\bf z} L$. Its job now is twofold. On the one hand, it computes the gradient with respect to its weights and bias,

\begin{align}\label{eq:grad_weights}
    \nabla_{\bf W} L &= (\nabla_{\bf z} L) \bf x ^T, \\
    \nabla_{\bf b} L &= \nabla_{\bf z} L,
\end{align}

\noindent so that it can be used to perform an update in the direction that minimizes $L$. On the other hand, it computes the gradient of $L$ with respect to the layer's input $\bf x$,

\begin{equation}\label{eq:grad_input}
    \nabla_{\bf{x}} L = \bf W^T (\nabla_{\bf z} L),
\end{equation}

\noindent to feed it to the backward pass of the previous layer, continuing the backpropagation algorithm.
The idea we propose is to modify the backward pass while keeping the forward pass the same.

\subsection{The BGN layer}
\label{sec:bbn}

In particular, we implement a new layer for backward gradient normalization. In the forward pass it simply computes an identity function of its input:

\begin{equation}
    \text{BGN}_{f} ({\bf x}) = {\bf x}.
\end{equation}

\noindent Therefore, it can be placed after any other layer without disturbing the results of its forward pass. 
In the backward pass, similarly to the standard layer discussed before, our layer receives a gradient ${\bf g}$, but instead of passing it backward untouched, it normalizes the gradient using the entire batch data (in fact all the dimensions of the tensor are used):

\begin{equation}
    \text{BGN}_{b} ({\bf g}) = \kappa \frac{{\bf g}}{||{\bf g}||},
\end{equation}

\noindent where $||\cdot||$ is the L$^2$ matrix norm and $\kappa$ is a scalar constant. In all our experiments we include a BGN layer in the network just before every activation function, and set $\kappa = \sqrt{d}$, with $d$ being the dimension of ${\bf g}$ excluding the batch size.

Let us take as an example the $K$-th layer in a deep neural network with activation function $f$ and weight matrix and bias ${\bf W}$ and ${\bf b}$, respectively. Its backpropagation equation, similarly to equation \ref{eq:grad_input}, is

\begin{equation}\label{eq:standard_backprop}
    \boldsymbol{\delta}^{(K)} = [{\bf W}^{T} \boldsymbol{\delta}^{(K+1)}] \circ f'({\bf z}^{(K)}),
\end{equation}

\noindent where $\boldsymbol{\delta}^{(K)}$ represents the gradient of the loss function with respect to the output ${\bf z}^{(K)}$ (before application of the non-linearity) of layer $K$ and $\circ$ is an element-wise tensor product. If we were now to introduce a BGN layer in between this layer's weighted sum and its activation function, as we pointed out earlier, its forward pass would not be affected, but equation \ref{eq:standard_backprop} would become

\begin{equation}\label{eq:bbn_backprop}
    \boldsymbol{\delta}_\text{BGN}^{(K)} = \kappa \cdot \frac{[{\bf W}^{T} \boldsymbol{\delta}_\text{BGN}^{(K+1)}] \circ f'({\bf z}^{(K)})}{||[{\bf W}^{T} \boldsymbol{\delta}_\text{BGN}^{(K+1)}] \circ f'({\bf z}^{(K)})||}.
\end{equation}

\noindent If we abbreviate the norm as $n_K = \frac{1}{\kappa} ||[{\bf W}^{T} \boldsymbol{\delta}_\text{BGN}^{(K+1)}] \circ f'({\bf z}^{(K)})||$, we get a shortened version of equation \ref{eq:bbn_backprop}:

\begin{equation}\label{eq:bbn_backprop_red}
    \boldsymbol{\delta}_\text{BGN}^{(K)} = \frac{1}{n_K}[{\bf W}^{T} \boldsymbol{\delta}_\text{BGN}^{(K+1)}] \circ f'({\bf z}^{(K)}).
\end{equation}

\noindent It is easy to see now through this recursion that for an $N$ layers deep neural network:

\begin{equation}
     \boldsymbol{\delta}_\text{BGN}^{(K)} = \boldsymbol{\delta}^{(K)} \prod_{j=K}^N \frac{1}{n_j} \quad \forall K \in {1,\dots,N}.
\end{equation}

\noindent This means that, for any given layer with backward gradient normalization, the backpropagation it receives is the same as in the case with no gradient normalization multiplied by a scalar. The gradients with respect to the weights and biases would then be:

\begin{align}\label{eq:weights_grad}
    \nabla_{\bf W} L &= \boldsymbol{\delta}_\text{BGN}^{(K)} \bf x^T, \\
    \nabla_{\bf b} L &= \boldsymbol{\delta}_\text{BGN}^{(K)}.
\end{align}

\noindent Therefore, the update to the weights and bias is proportional to the one in the case without BGN, but with a different proportion for each layer. This would be equivalent to assigning a different learning rate for every layer \cite{DBLP:journals/corr/YuLSC17}.

In a way, we could consider $\boldsymbol{\delta}_\text{BGN}^{(K)}$ to have norm $\kappa$, so $\nabla_{\bf W} L$ would have the same norm as $x$, the input to that layer (or the output of the previous one), scaled by $\kappa$. This, in turn, avoids the vanishing and exploding gradients problems, as it ensures that the gradient of each layer is approximately as big as its input, no matter how deep the network is.

\section{Visualizing the gradients}
\label{sec:gradient_visualization}

In order to understand how the gradients backpropagate through the network, we show in figure \ref{fig:gradients_90_init} a plot of the gradient norms in a neural network with $90$ hidden layers and different training configurations. We consider networks with three different types of activation function, namely sigmoid, tanh and ReLU. And we include both batch normalization \cite{DBLP:conf/icml/IoffeS15} and the backward gradient normalization technique introduced in section \ref{sec:gradnorm} as methods to control the vanishing/exploding gradients problems. The remaining network parameters are set as described in section \ref{sec:experiments}. 

\subsection{Initial gradients}
\label{sec:initial_gradients}

\begin{figure}[htb]
\begin{center}
\includegraphics[width=0.32\textwidth]{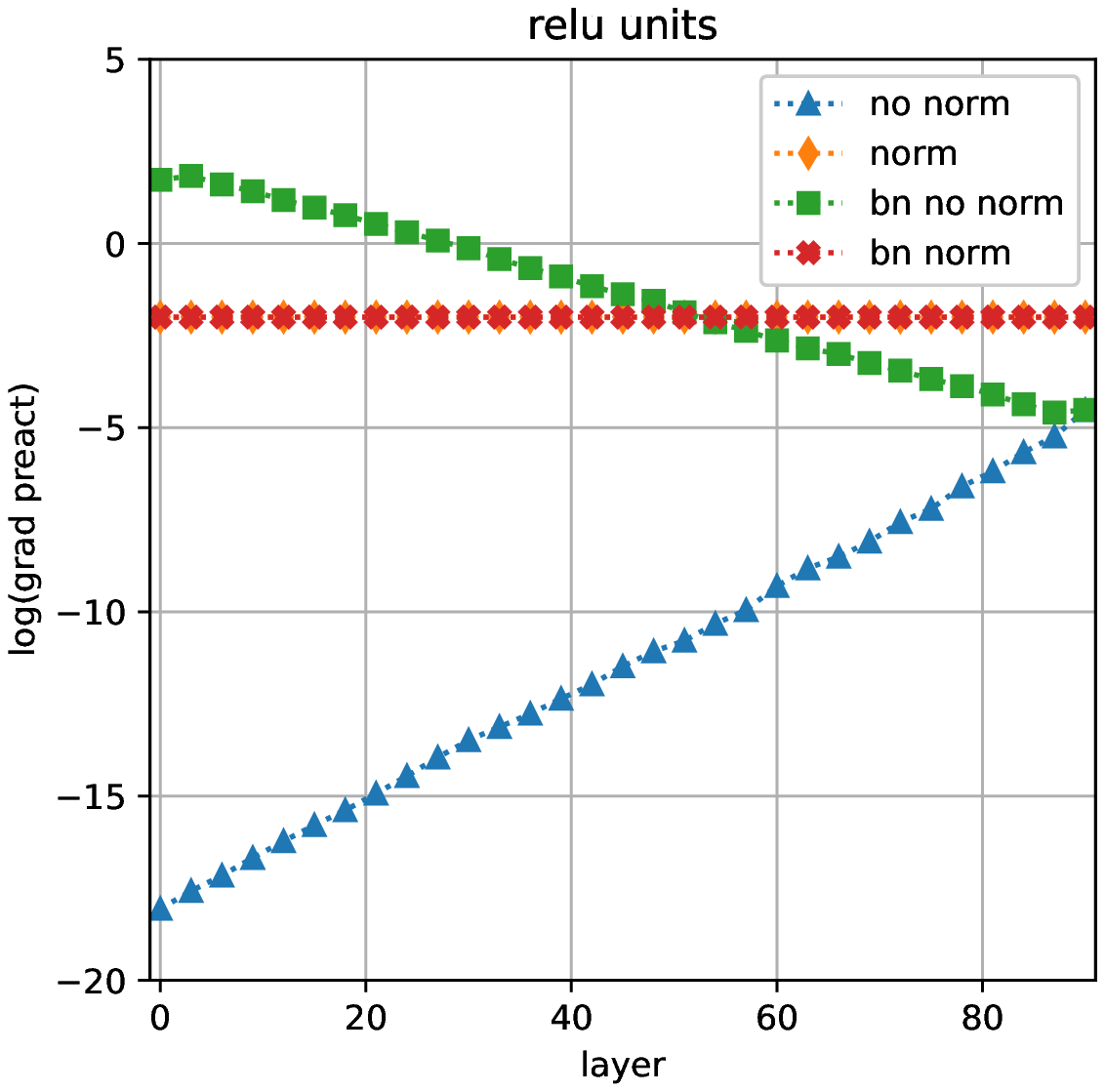}
\includegraphics[width=0.32\textwidth]{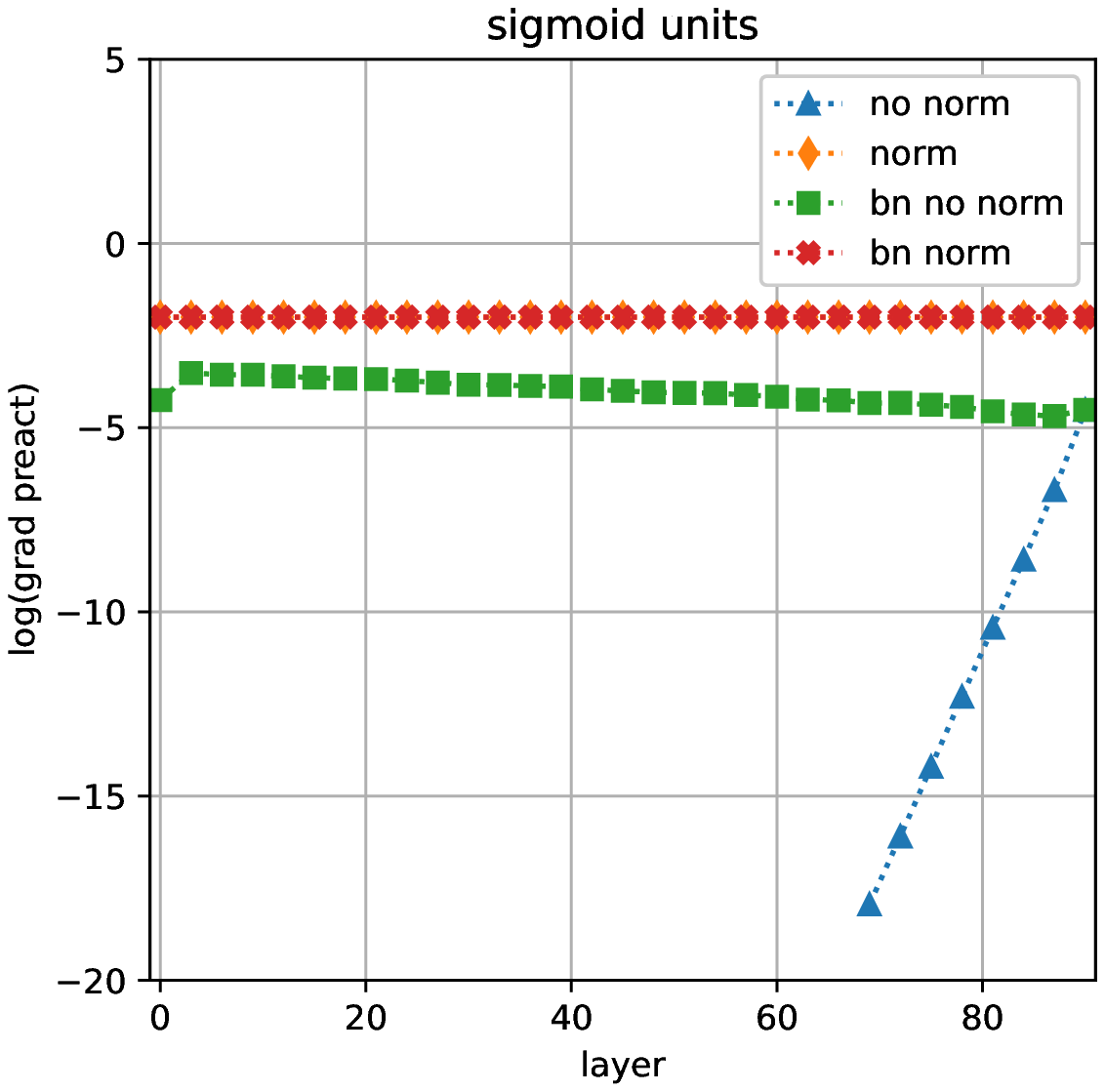}
\includegraphics[width=0.32\textwidth]{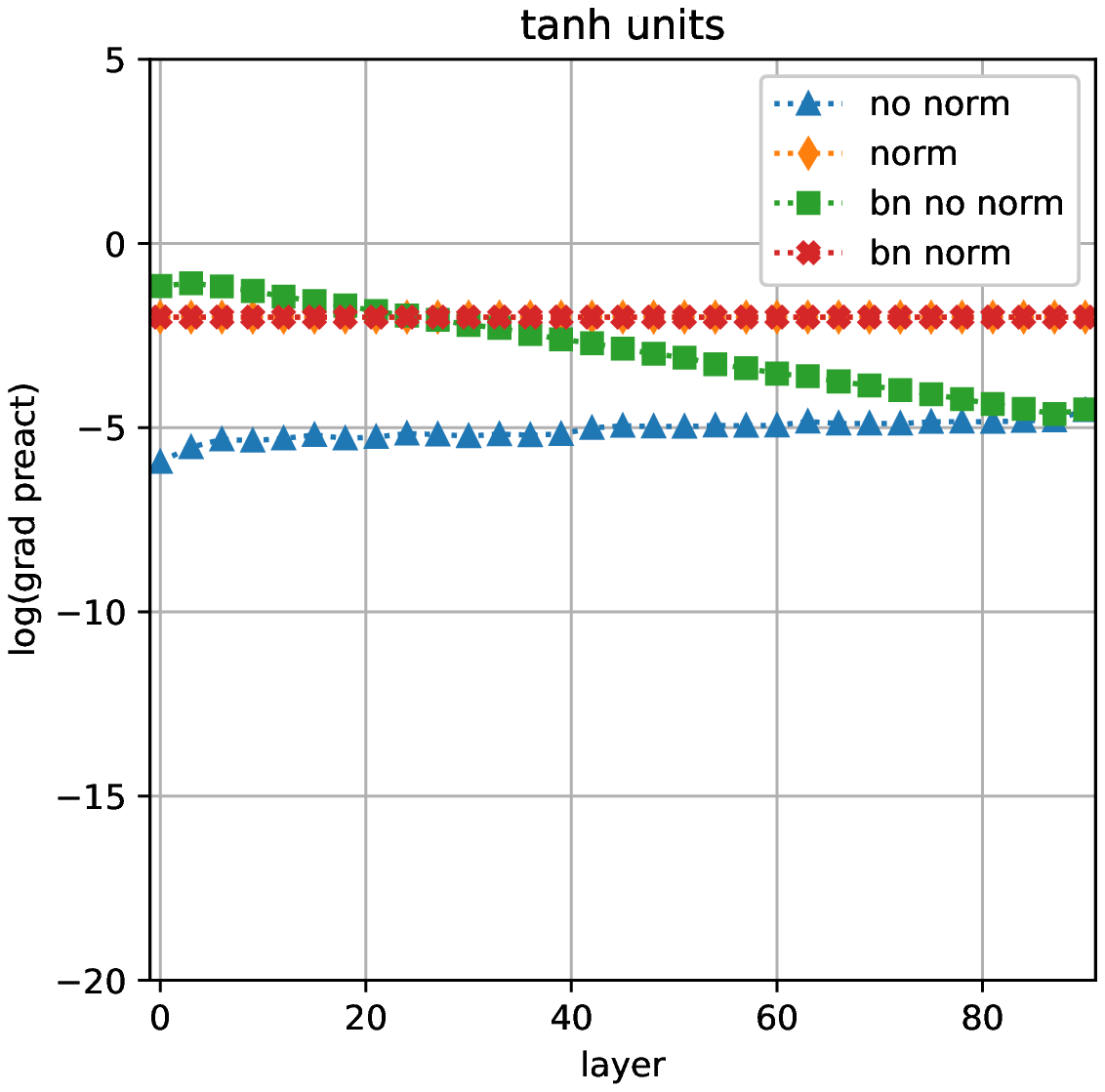}\\
\includegraphics[width=0.32\textwidth]{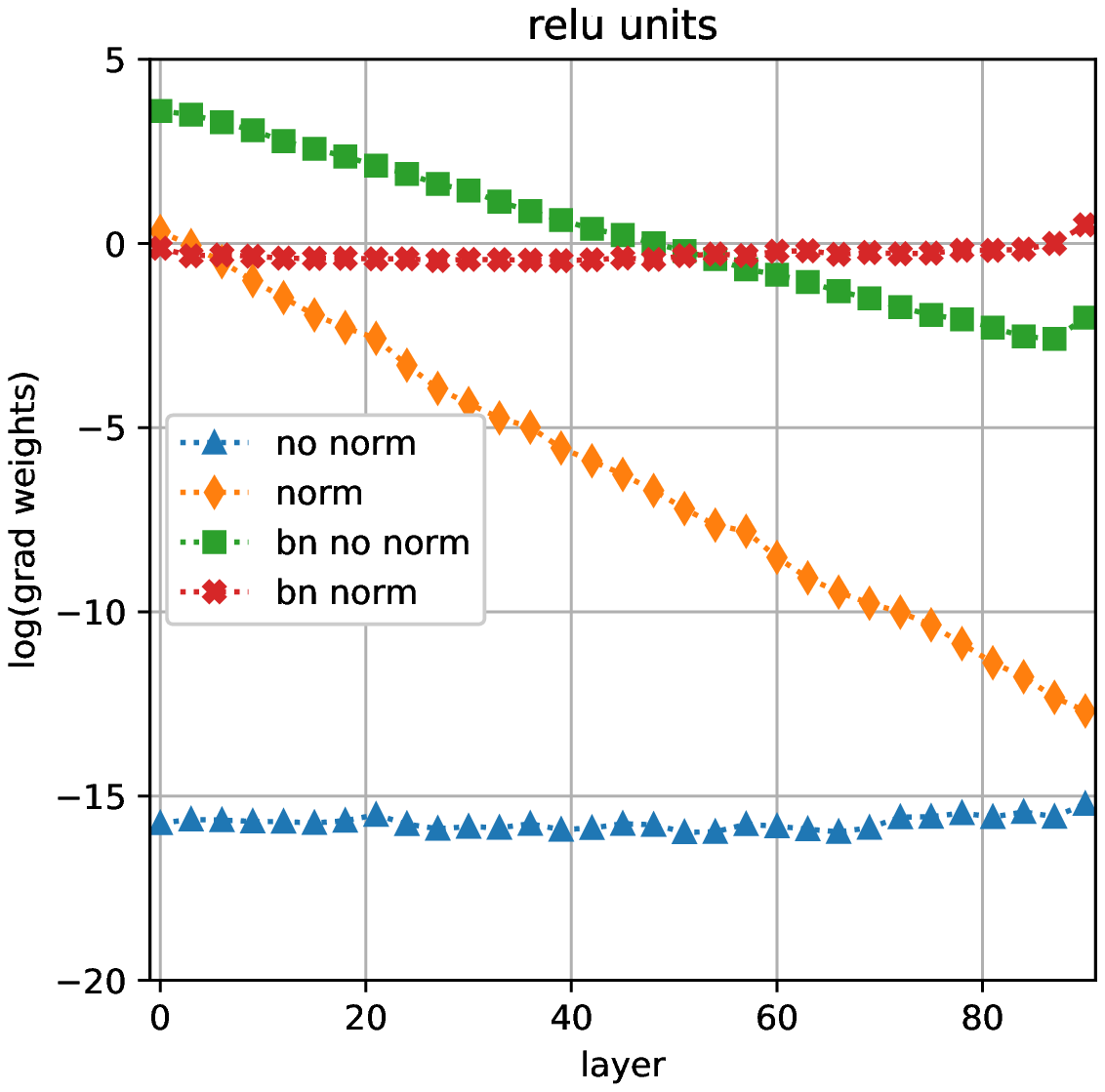}
\includegraphics[width=0.32\textwidth]{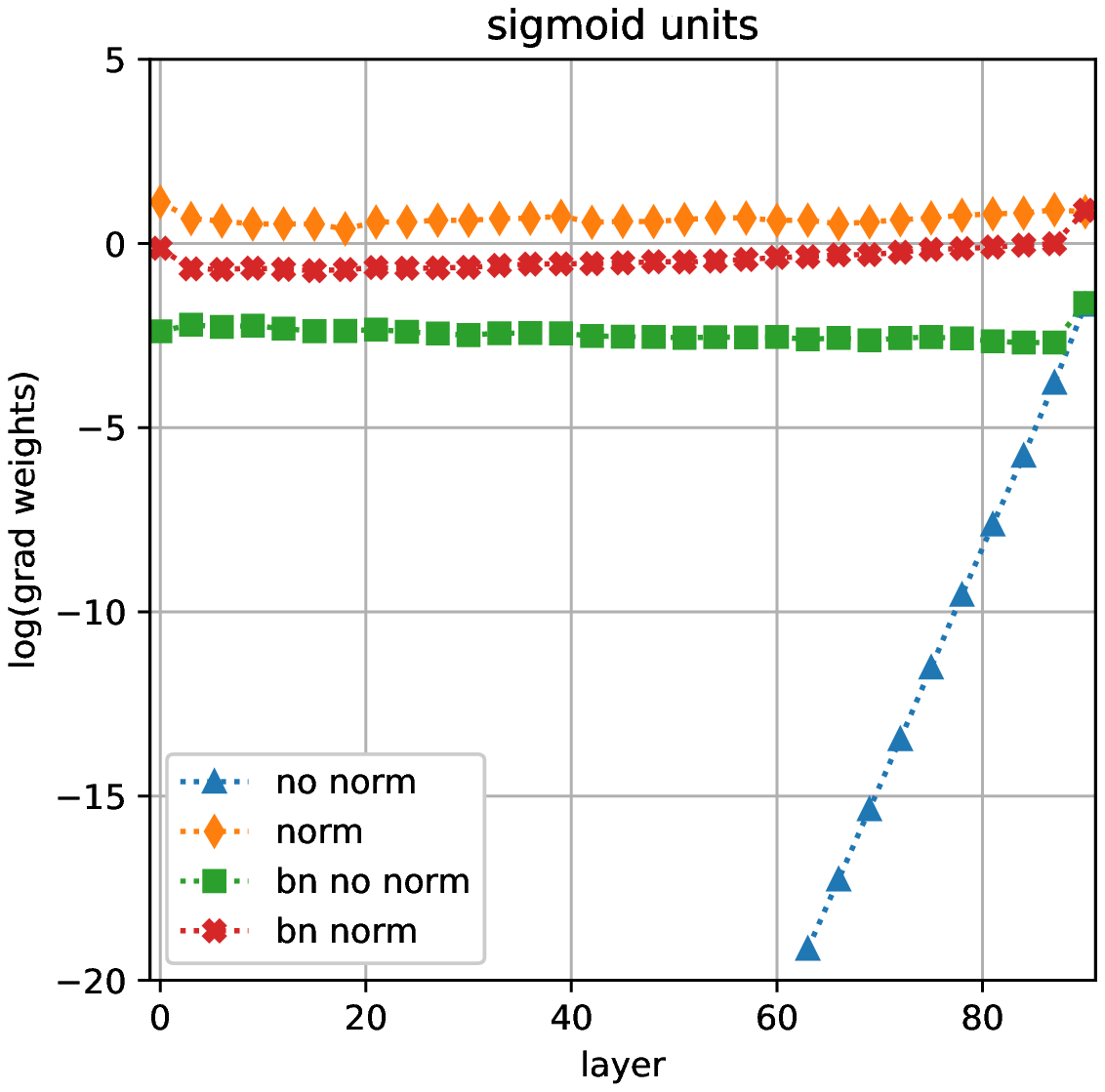}
\includegraphics[width=0.32\textwidth]{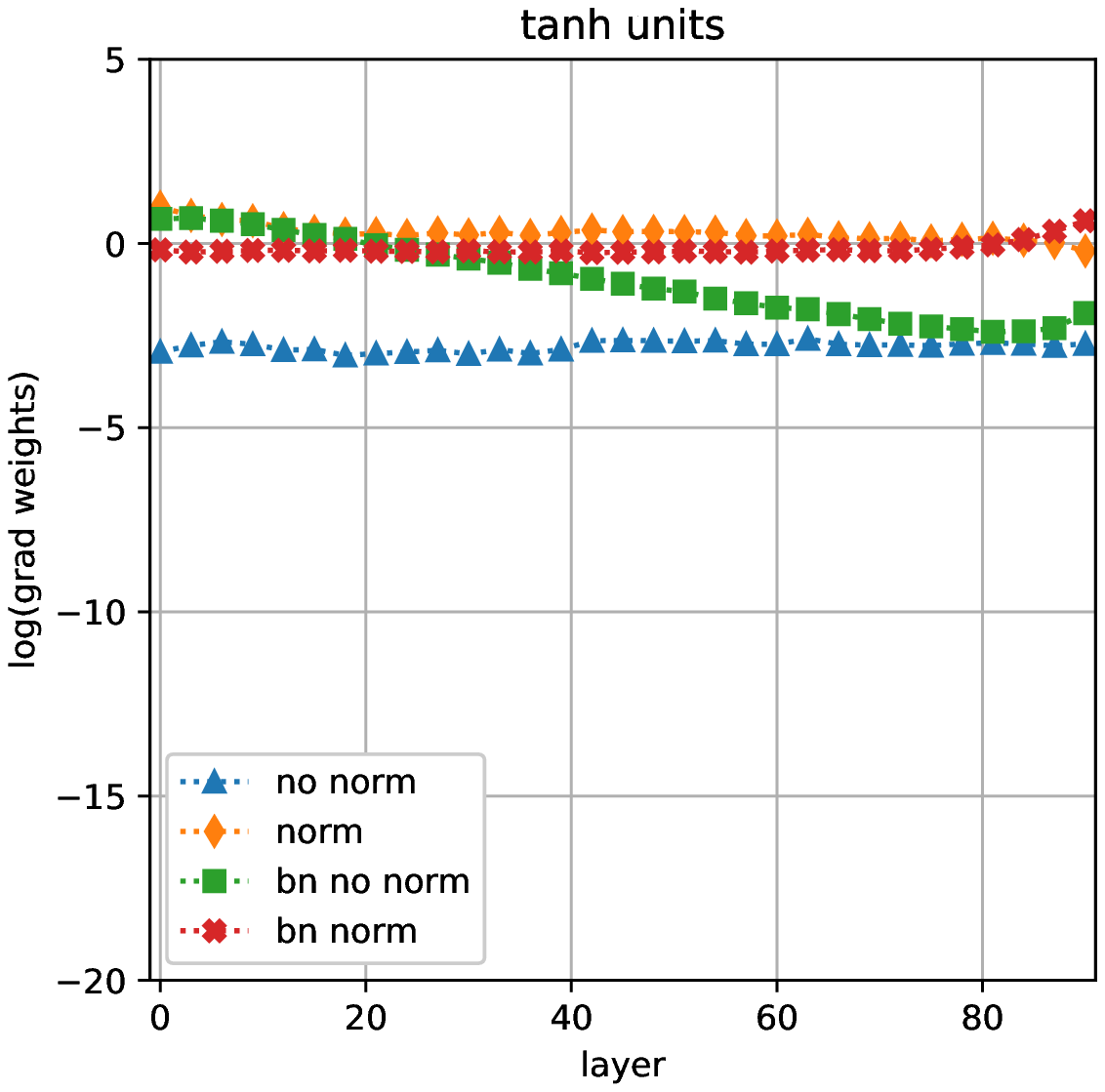}
\end{center}
\caption{Initial gradients, after weight initialization, in a neural network with 90 layers.} 
\label{fig:gradients_90_init}
\end{figure}

The top plots in figure \ref{fig:gradients_90_init} show the gradient norms ($||\boldsymbol{\delta}^{(K)}||$ or $||\boldsymbol{\delta}_{BGN}^{(K)}||$) when the gradients of the loss function are computed with respect to each layer's pre-activation (equations \ref{eq:standard_backprop} and \ref{eq:bbn_backprop}). The bottom plots are equivalent, but the gradients are computed with respect to the weight matrices. Each column refers to a different activation: ReLU (left), sigmoid (middle) and tanh (right). When no kind of normalization is applied (blue plots), we observe an exponential decay of  $||\boldsymbol{\delta}^{(K)}||$ with the layer depth (note the logarithmic scale in the $y$ axis). It is specially significant for the sigmoid activation, and seems to affect the tanh units to a lesser extent. When including BGN, however, the gradient norms $||\boldsymbol{\delta}_{BGN}^{(K)}||$ are constant and independent of the layer depth in all cases (yellow plots). 

The gradients with respect to the network weights depend on the preactivation gradients $\boldsymbol{\delta}^{(K)}$, but also on the layer inputs. This dependence affects each activation function in a different manner (figure \ref{fig:gradients_90_init}, bottom plots). While the effect observed for the sigmoid and tanh units is mapped almost unchanged from $\boldsymbol{\delta}^{(K)}$ to $\bf W$, the ReLU neurons can compensate the gradient decay and $||\nabla_{\bf W} L||$ is almost constant (although very small) for all the network layers. The use of gradient normalization seems to be beneficial for both the sigmoid and the tanh, while its effect on the ReLU is more uncertain. 
How all these observations affect network learning, and the effect on the prediction accuracy, is however unclear and will be analyzed in later sections. The observed behavior can change if we consider other types of normalization, such as batch normalization.

\subsection{The effect of batch normalization}
\label{sec:batch_normalization}

Batch normalization \cite{DBLP:conf/icml/IoffeS15} can improve network learning by normalizing the activations before applying the activation function. So we expect that it affects the gradients as well. The gradient norms computed using BN are also shown in figure \ref{fig:gradients_90_init}. The green plots illustrate the gradient norms when BN is used alone, while the red plots are for the combined use of BN and BGN. Batch normalization affects $\boldsymbol{\delta}^{(K)}$ by increasing the gradient norm with the layer depth. This effect is more pronounced for ReLU and tanh units, and almost imperceptible for the sigmoid. When BGN and BN are used in conjunction, the plots flatten and the gradient norms are almost constant for all depths. The same observations apply for the gradients with respect to the layers' weights, which follow closely the behavior of $\boldsymbol{\delta}^{(K)}$. It is interesting to observe that when $\bf{x}$ is (batch) normalized, the gradients with respect to $\bf{W}$ are basically $\boldsymbol{\delta}^{(K)}$.

All the gradients in figure \ref{fig:gradients_90_init} have been computed just after network initialization. Standard weight initialization techniques are designed to favor the initial propagation of gradients. Hence we can expect that the vanishing and/or exploding gradients problems are initially underestimated. Weight adaptation after network learning could place the networks in a worse situation. In the next section we repeat the former analysis after network training.   

\subsection{Gradient visualization after network training}
\label{sec:after_training}

\begin{figure}[htb]
\begin{center}
\includegraphics[width=0.32\textwidth]{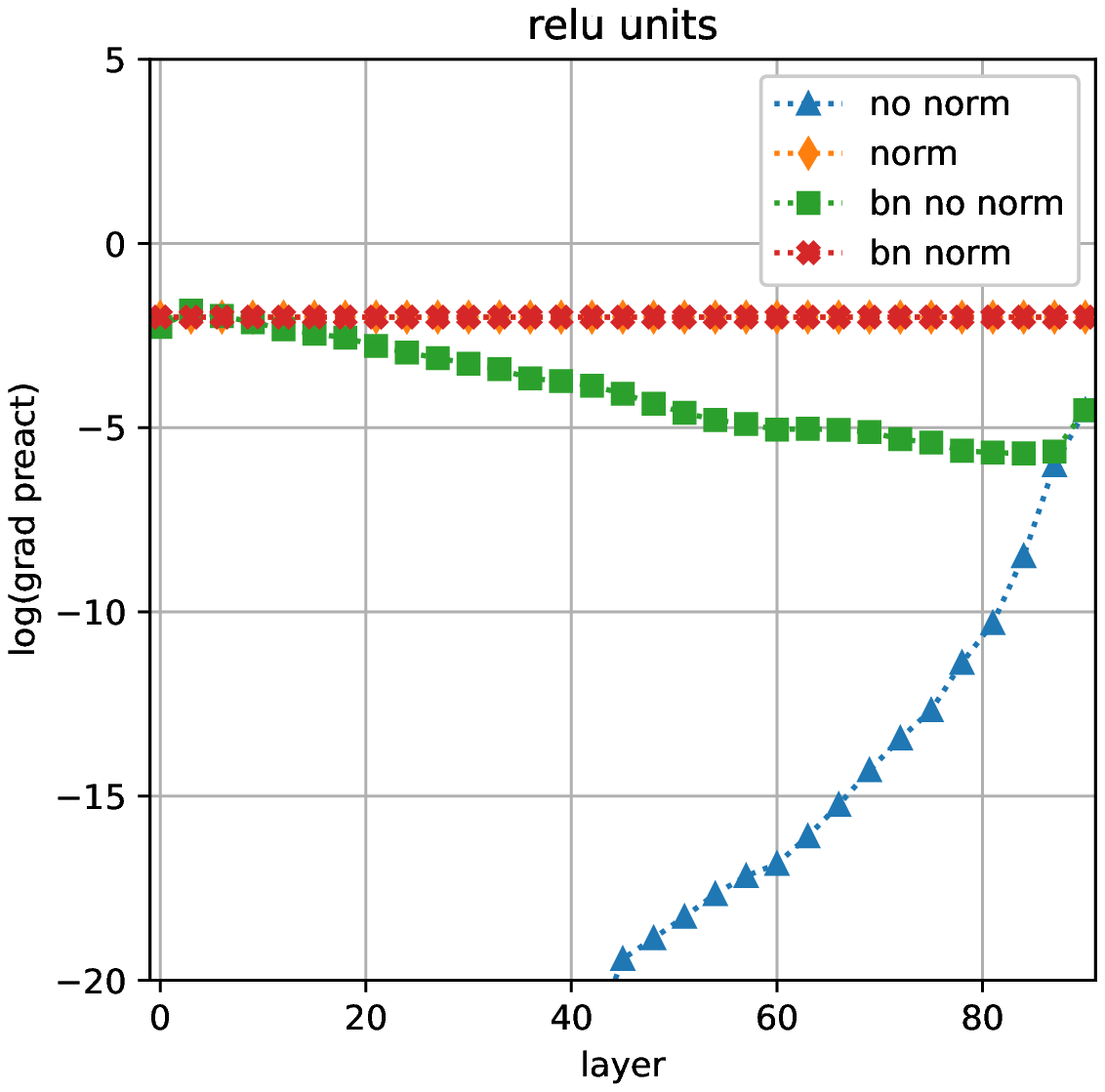}
\includegraphics[width=0.32\textwidth]{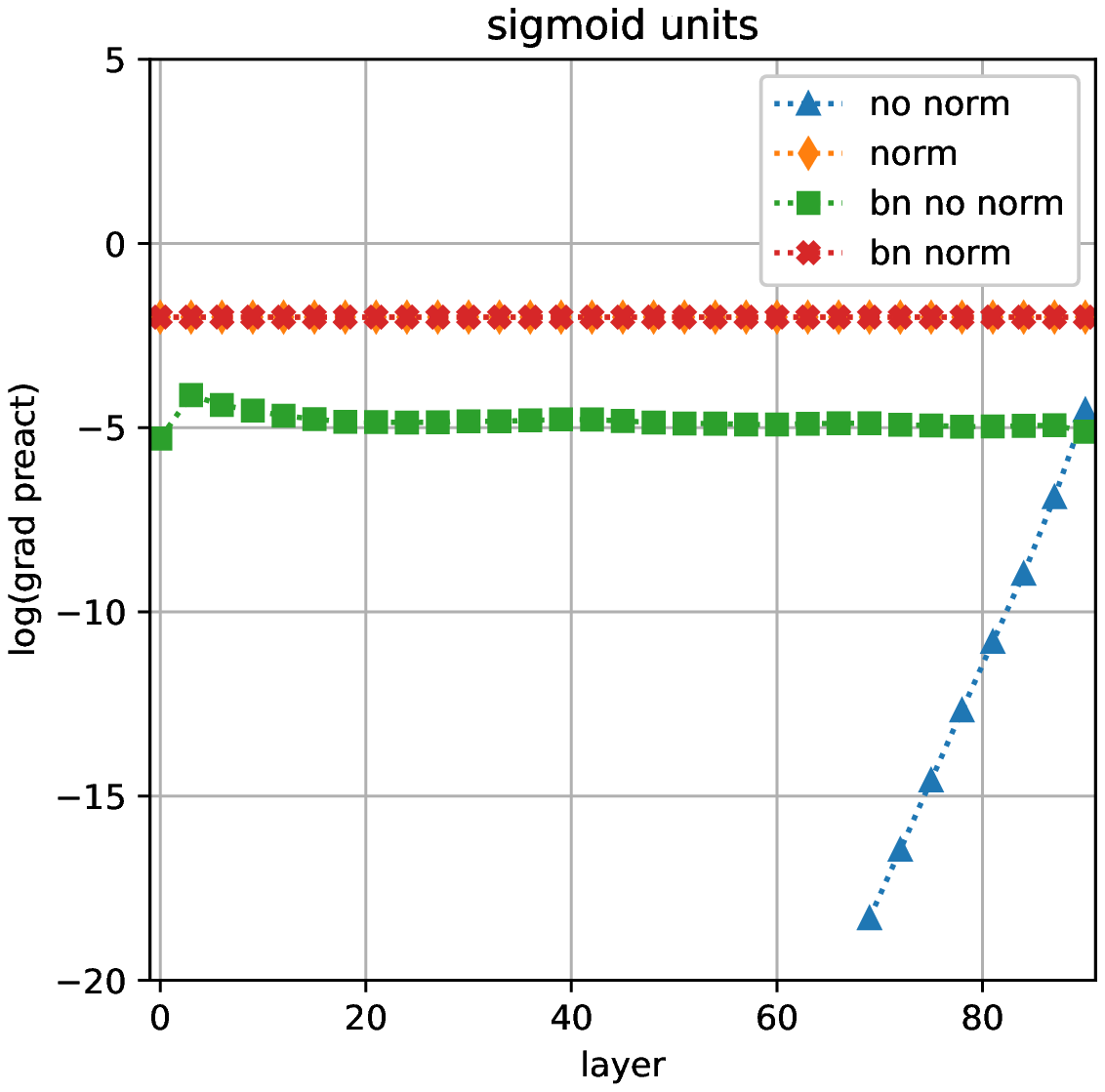}
\includegraphics[width=0.32\textwidth]{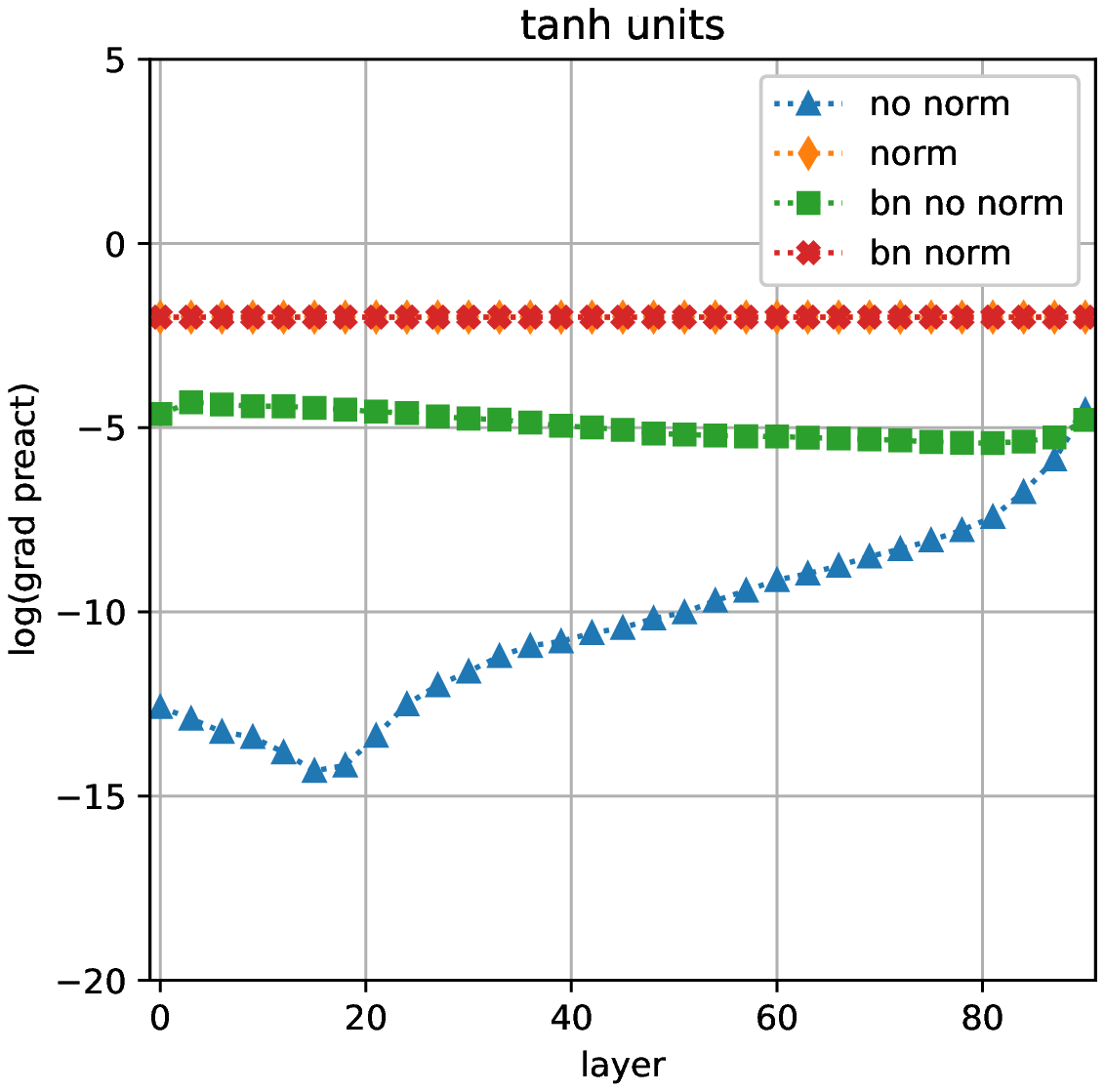}\\
\includegraphics[width=0.32\textwidth]{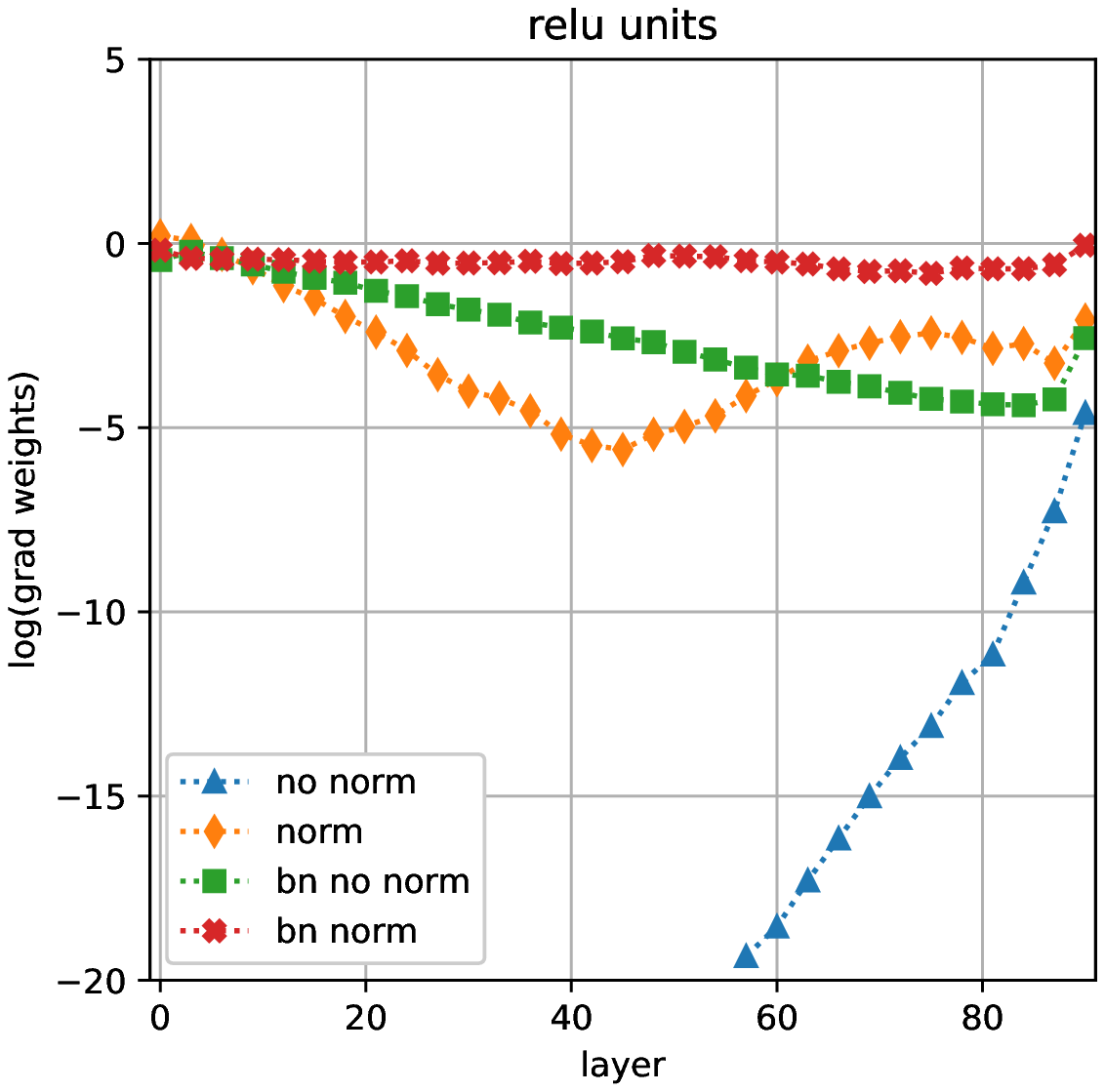}
\includegraphics[width=0.32\textwidth]{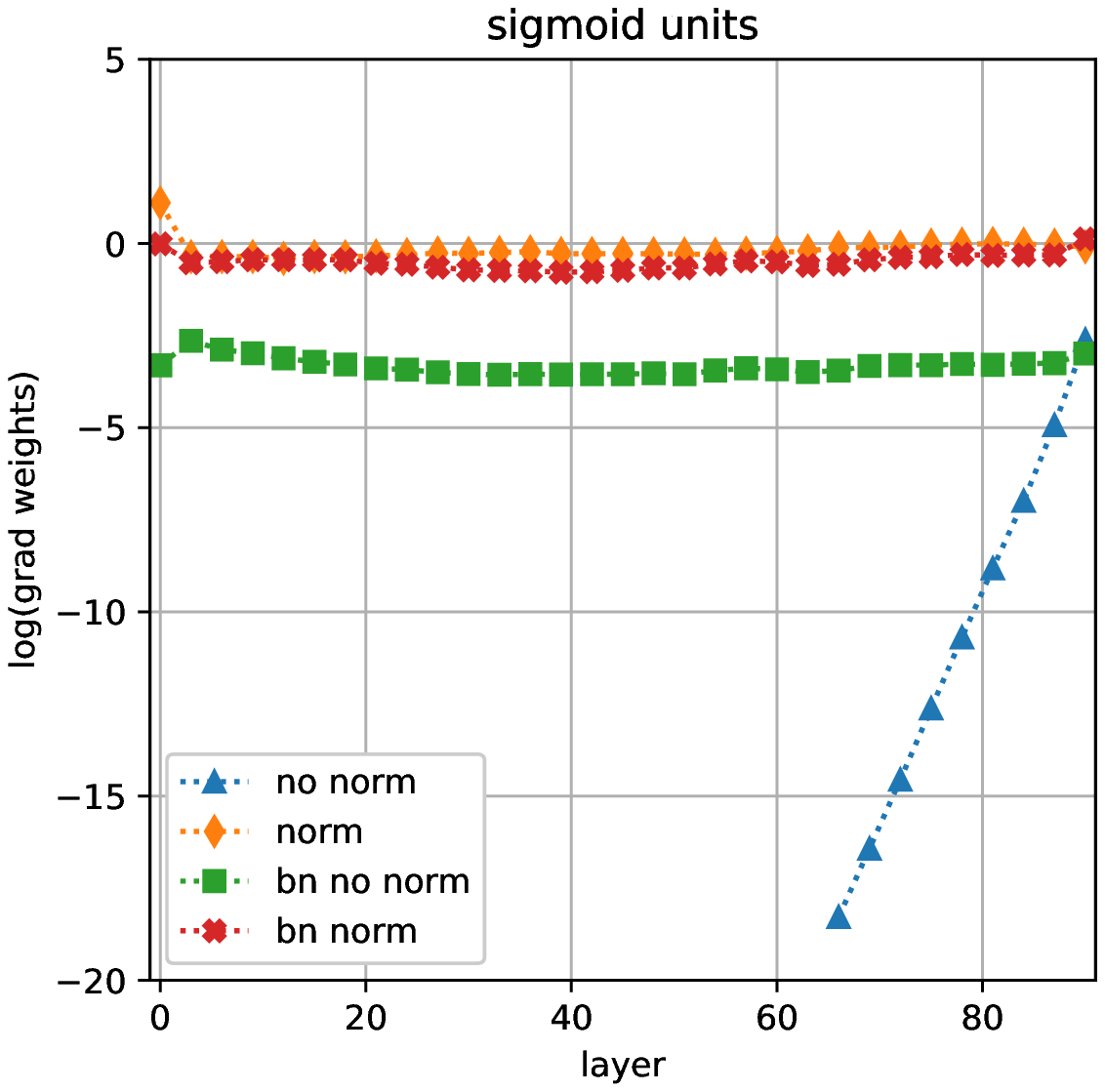}
\includegraphics[width=0.32\textwidth]{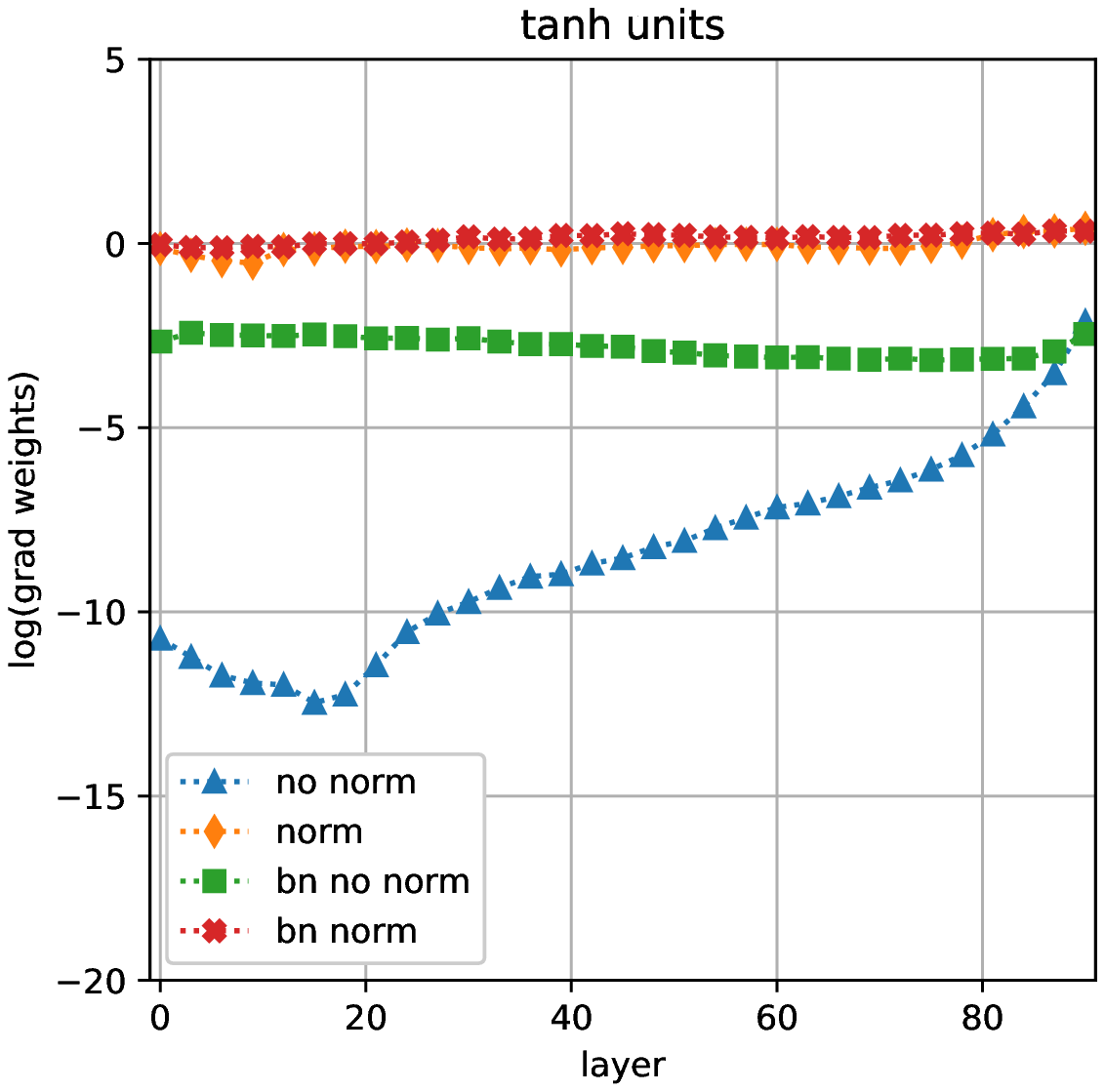}
\end{center}
\caption{Gradients after the networks have been trained for 20 epochs.} \label{fig:gradients_90_after_training}
\end{figure}

Departing from the networks described in sections \ref{sec:initial_gradients} and \ref{sec:batch_normalization}, we have trained them for 20 epochs (see experimental details below) and observed the gradients after training. All the networks are trained without BGN. The results are shown in figure \ref{fig:gradients_90_after_training}. The main difference between the two figures is that the vanishing gradients problem is now more evident when no kind of normalization is used (blue plots). It affects the networks regardless of the activation type, although it is less striking for the tanh. BGN seems to correct this effect for the sigmoid and tanh activations and, to a lesser extent, also for the ReLU neurons (yellow plots).
The use of BN alone also helps to correct the gradients decay (green plot). Finally, the combination of batch normalization and BGN (red plots) seems to be the most stable option in all cases, with the gradient norms being almost constant and close to 1 for all the network layers. 

These results show that the BGN technique is effective for preventing the gradient from vanishing or exploding. However that is not a guarantee of a good performance when the networks are trained to solve a problem. In the next section we perform a comparison of all the previous strategies on a classification problem.

\section{Experiments}
\label{sec:experiments}

Although the analysis of section \ref{sec:gradient_visualization} can help to gain intuition on how the different normalization techniques affect the gradient propagation in a neural network, there are many other processes involved during network training. How all these processes interact to determine the final network weights, and hence its prediction accuracy, can be quite intricate and difficult to understand in terms of simple principles. This is why the final advantages of a new algorithm or technique must be evaluated by measuring its performance on benchmark problems. In this section we describe a set of experiments designed to compare the prediction accuracy of neural networks trained under several of the conditions described above.

All the experiments are performed using the permutation invariant MNIST dataset. As our goal is to observe how BGN affects gradients in deep neural networks, we chose to use simple dense networks with depths of 30, 60, 90 and 120 layers, each of which composed of 64 units. In order to compare several scenarios, we ran each architecture with every combination of the following hyperparameters: ReLU, sigmoid and tanh activations, with and without batch normalization and, of course, with or without BGN layers just before every activation function. All of these models are trained using the Adam optimizer \cite{DBLP:journals/corr/KingmaB14}.

After an initial exploration, we trained each of these models with 19 learning rates between $10^{-4}$ and $10^{-2}$, for 20 epochs. We repeated this 5 times for each combination, in order to get a mean value of accuracy for each learning rate. Knowing the learning rate with the highest average accuracy for each combination of hyperparameters, we trained those 10 more times each for 20 epochs, ending up in total with 15 values of accuracy for each combination. We calculated the mean and standard deviation, and measured the running times. The results are presented in section \ref{sec:results}.

\section{Results}
\label{sec:results}

From the accuracy metrics displayed in table \ref{table:results} alone, we can already reach some conclusions. For starters, for almost every activation function and number of layers, the highest accuracy is reached when we use BGN; and in the cases where it is not, the corresponding results with BGN are not far behind. Also, in most cases, the accuracy reached through BGN is higher than the corresponding non-BGN counterpart, so the addition of BGN pretty much always improves the results.

\begin{table}[]
    \small
    \centering
    \caption{Mean $\pm$ standard deviation of accuracy for 15 independent 20-epoch-long trainings, for every combination of three activation functions, with or without batch normalization, with or without BGN and four different depths. The highest mean accuracy per activation function and depth is highlighted in bold font.}
        \begin{tabular}{ |l|c|c|c|c|c|c| } 
        \hline
        activation & BN & BGN & 30 layers & 60 layers & 90 layers & 120 layers\\ 
        \hline
    relu &    False &    False & $0.948 \pm 0.006$ & $0.729 \pm 0.319$ & $0.114 \pm 0.000$ & $0.114 \pm 0.000$ \\
    relu &    False &     True & $0.949 \pm 0.007$ & $\bf{0.890} \pm 0.098$ & $\bf{0.901} \pm 0.039$ & $\bf{0.858} \pm 0.045$ \\
    relu &     True &    False & $0.958 \pm 0.002$ & $0.405 \pm 0.058$ & $0.141 \pm 0.026$ & $0.115 \pm 0.001$ \\
    relu &     True &     True & $\bf{0.968} \pm 0.002$ & $0.637 \pm 0.135$ & $0.159 \pm 0.038$ & $0.112 \pm 0.026$ \\
    \hline
 sigmoid &    False &    False & $0.114 \pm 0.000$ & $0.114 \pm 0.000$ & $0.113 \pm 0.002$ & $0.114 \pm 0.000$ \\
 sigmoid &    False &     True & $0.206 \pm 0.003$ & $0.200 \pm 0.025$ & $0.169 \pm 0.045$ & $0.181 \pm 0.038$ \\
 sigmoid &     True &    False & $\bf{0.961} \pm 0.003$ & $\bf{0.954} \pm 0.004$ & $\bf{0.949} \pm 0.004$ & $0.940 \pm 0.010$ \\
 sigmoid &     True &     True & $0.958 \pm 0.001$ & $0.952 \pm 0.006$ & $0.947 \pm 0.005$ & $\bf{0.945} \pm 0.005$ \\
    \hline
    tanh &    False &    False & $0.956 \pm 0.002$ & $0.944 \pm 0.003$ & $\bf{0.935} \pm 0.005$ & $0.896 \pm 0.017$ \\
    tanh &    False &     True & $0.954 \pm 0.002$ & $0.948 \pm 0.003$ & $0.928 \pm 0.028$ & $\bf{0.901} \pm 0.013$ \\
    tanh &     True &    False & $0.963 \pm 0.003$ & $0.936 \pm 0.003$ & $0.870 \pm 0.025$ & $0.516 \pm 0.076$ \\
    tanh &     True &     True & $\bf{0.964} \pm 0.002$ & $\bf{0.953} \pm 0.002$ & $0.917 \pm 0.009$ & $0.758 \pm 0.126$ \\
    \hline
    \end{tabular}
    \label{table:results}
    \normalsize
\end{table}

If we focus on each activation function, we can also see some patterns. For instance, with ReLU, the use of batch normalization does nothing to help learning in networks deeper than 30 layers, whereas BGN alone achieves similar results for 30 layers and manages to learn the problem with deeper networks. For the sigmoid activations, however, the roles are reversed. BGN gets better results than the standard network, but still does not learn the problem. Meanwhile, both cases with batch normalization (with and without BGN) achieve very similar high accuracy. It is interesting to note that the performance of sigmoid based networks is almost invariant with the number of layers when batch normalization is used, independently of whether BGN is used or not. In the case of tanh, it does not seem to be affected too much by either batch normalization or BGN; in this sense it is a robust activation function. However, with 90-layer-deep networks, batch normalization starts to lower the performance and, even though BGN helps compensate it, this effect becomes clear in the case of 120 layers.

As for training times, they are approximately the same across all three activation functions and grow proportionally to the number of layers. It is important to note that, while batch normalization almost triples the time required for training in every setting ($ 40.00 \pm 0.31$s vs. $118.00 \pm 0.82$s with 90 layers, ReLU, no BGN; $ 50.83 \pm 2.24$s vs. $127.78 \pm 0.90$s with BGN), BGN just adds around 10 seconds to the training time ($ 40.00 \pm 0.31$s vs. $ 50.83 \pm 2.24$s; $118.00 \pm 0.82$s vs. $127.78 \pm 0.90$s), which can make BGN more desirable in the cases where both strategies are comparable.

\begin{figure}[htb]
\begin{center}
\includegraphics[width=0.32\textwidth]{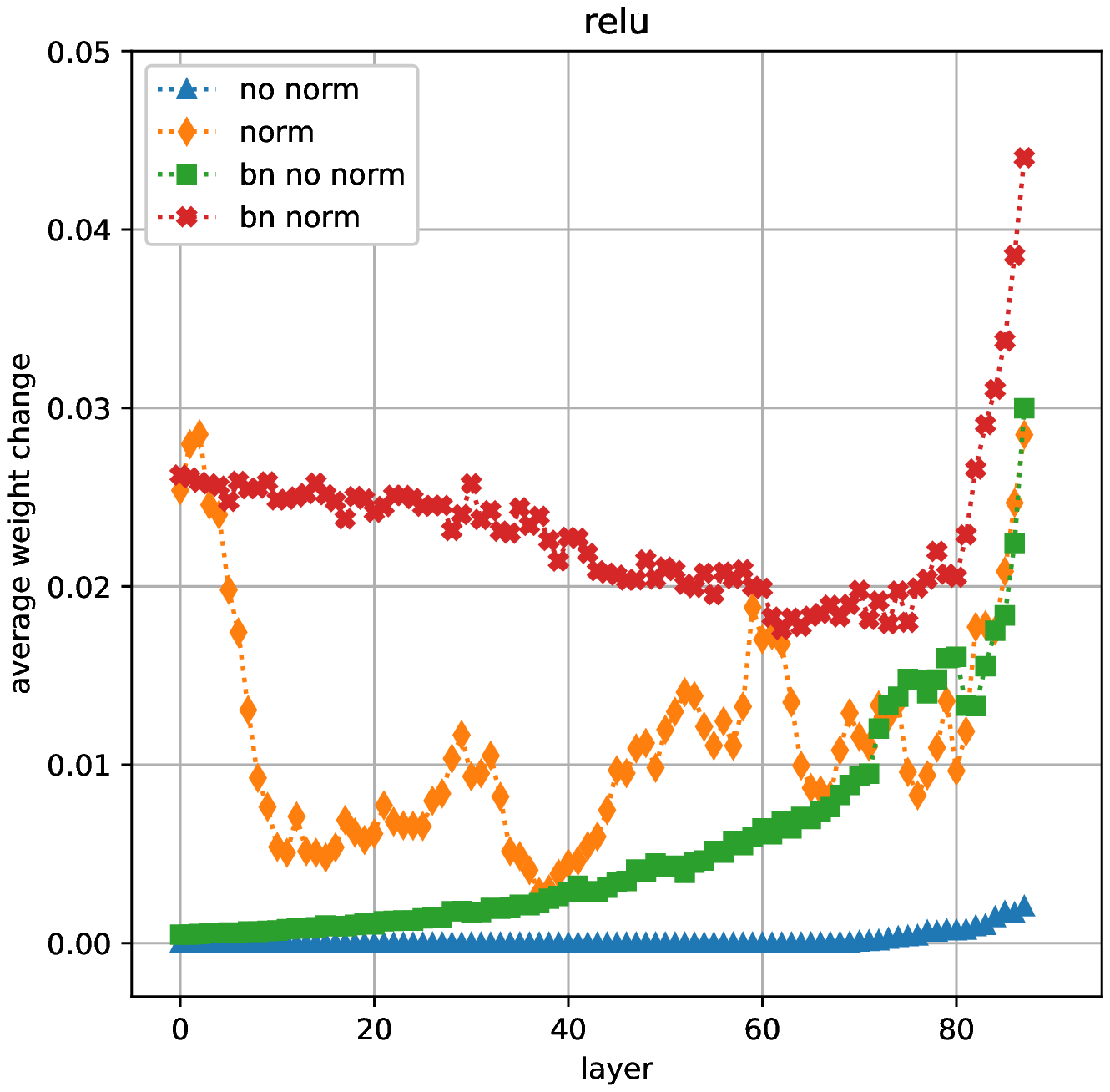}
\includegraphics[width=0.32\textwidth]{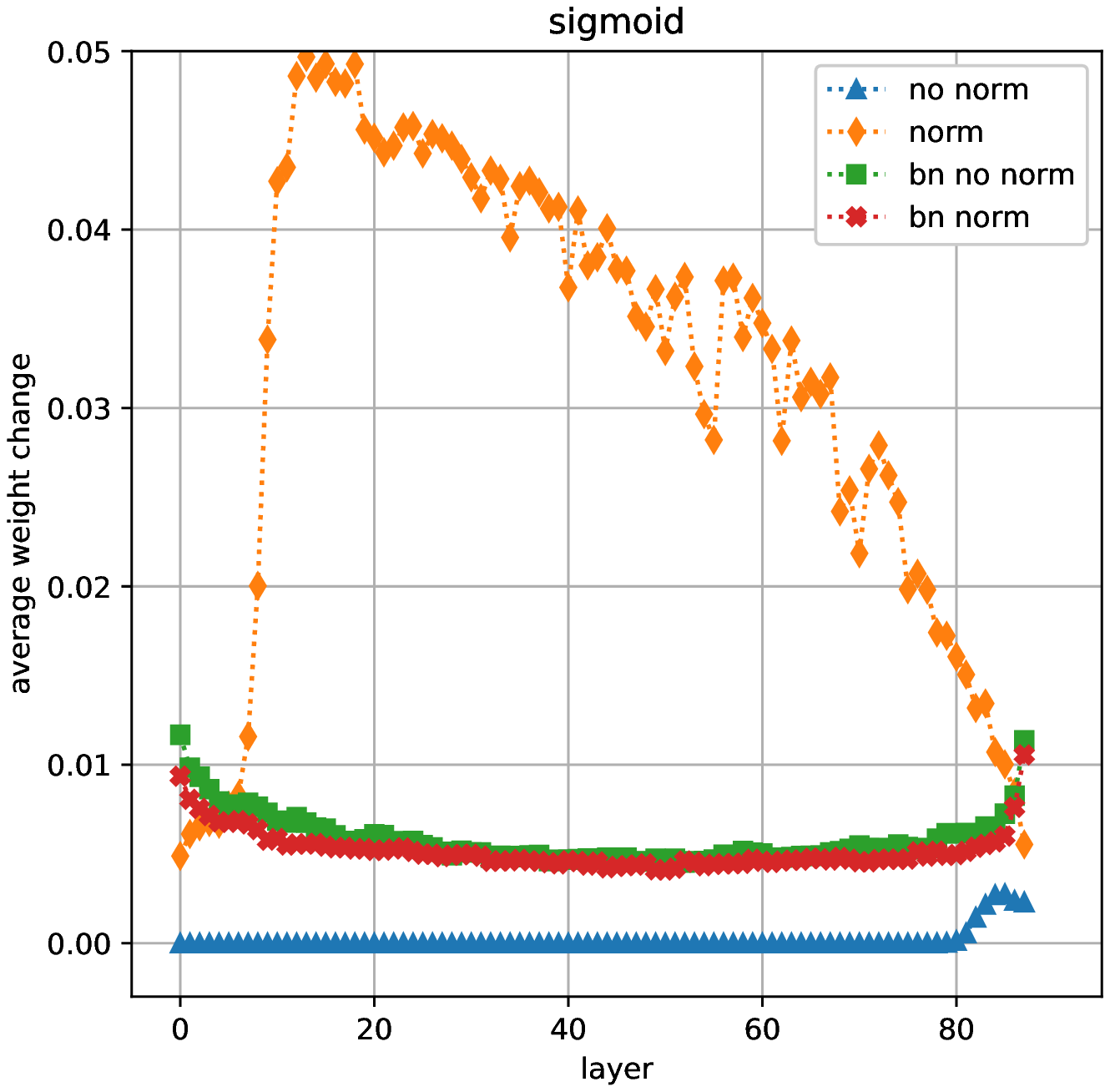}
\includegraphics[width=0.32\textwidth]{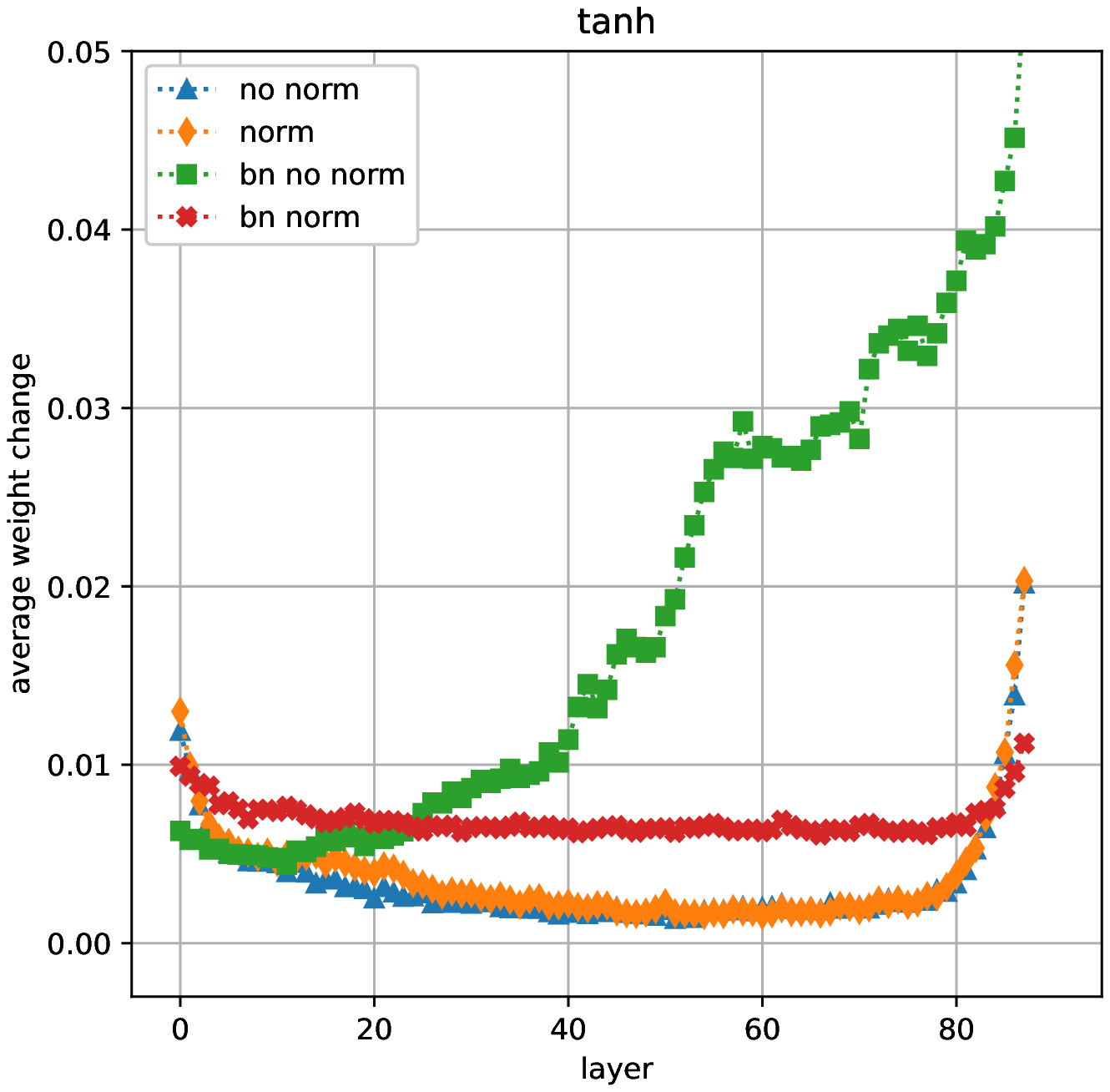}
\end{center}
\caption{Weight change after 20 training epochs in networks with 90 layers with ReLU activation (left), sigmoid activation (middle) and tanh activation (right). The data for the sigmoid activation with BGN have been rescaled (with a factor 0.08) to fit the plot.} 
\label{fig:weights}
\end{figure}

Finally, to complement previous results, we show in figure \ref{fig:weights} the average change in the network weights after training, for each layer in the networks of depth 90. The main observation is that the presence of BGN (red and yellow plots) favors the adaptation of all the network's weights in a similar scale. On the contrary, when the gradients are not normalized, there is in general a monotonic decrease in the adaptation level as the layer depth increases (with some nuances for each activation). Hence, the use of BGN not only improves the network's prediction accuracy, but also benefits a better distributed learning.

\section{Conclusions}
\label{sec:conclusions}

Several normalization techniques have been proposed for dealing with the vanishing and exploding gradients problems, with batch normalization \cite{DBLP:conf/icml/IoffeS15} being probably the most extended. In this article we have presented a new method for normalizing the gradients during the backward pass. It is similar to the layer-wise gradient normalization proposed in \cite{DBLP:journals/corr/YuLSC17}, but we directly backpropagate the normalized gradients. The resulting BGN technique permits a well-scaled gradient flow that reaches the deepest network layers without experimenting vanishing or explosion. The normalized gradients are proportional to the original gradients layer-wise. Hence the weight update direction is conserved, and the method can be understood as a standard gradient descent where different learning rates are used for different network layers. 

We have run several experiments to assess the ability of BGN to prevent the gradients from vanishing/exploding, measuring the prediction accuracy of networks of increasing depth, and comparing the new method to standard batch normalization. We show that the use of BGN is in general beneficial, specially for networks with ReLU neurons. In most of our tests the accuracy obtained with BGN is higher than the corresponding non-BGN counterpart, both with and without batch normalization. In addition, the training times of BGN networks are significantly reduced with respect to networks that use batch normalization. 

The problem data and the experimental settings have been chosen to illustrate the main claims of this work. In particular, we wanted to test the different normalization methods with networks of arbitrarily high depth. Although it could be argued that the network design is not very realistic given the selected problem, it allowed to demonstrate that BGN does an effective control of the gradients, obtaining accuracy levels that in the worst case are comparable to more standard approaches. While more systematic comparisons to state of the art methods on standard benchmark problems are still needed to validate the proposed technique, the results here presented are promising and open the door to improving  very deep neural networks training by gradient normalization.

\section*{Acknowledgement}

This work has been partially funded by grant S2017/BMD-3688 from Comunidad de Madrid and by Spanish projects MINECO/FEDER
TIN2017-84452-R and PID2020-114867RB-I00. AC acknowledges financial support from UAM (\textit{Ayudas para el fomento de la investigación en estudios de Máster}).

%
%
%
\bibliography{bbn}

\end{document}